\documentclass{article}
\usepackage{ijcai24}

\usepackage{times}
\usepackage{soul}
\usepackage{url}
\usepackage[hidelinks]{hyperref}
\usepackage[utf8]{inputenc}
\usepackage[small]{caption}
\usepackage{graphicx}
\usepackage{amsmath}
\usepackage{amsthm}
\usepackage{amsmath}
\usepackage{amsfonts}
\usepackage{booktabs}
\usepackage{algorithm}
\usepackage{algorithmic}
\usepackage[switch]{lineno}

\usepackage{xcolor}
\usepackage{tabularx}
\usepackage{multirow}
\newcommand{\raha}[1]{\textcolor{blue}{#1}}

\theoremstyle{definition}

\title{Causal Feature Selection for Responsible Machine Learning}

\author{
    Author Name
    \affiliations
    Affiliation
    \emails
    email@example.com
}

\author{
Raha Moraffah
\and
Paras Sheth$^\star$\and
Saketh Vishnubhatla\thanks{Equal Contribution}\And
Huan Liu\\
\affiliations
School of Computing and Augmented Intelligence, Arizona State University, USA\\
\emails
\{raha.moraffah, psheth5, svishnu6, huanliu\}@asu.edu
}

\begin{document}

\maketitle

\begin{abstract}
   Machine Learning (ML) has become an integral aspect of many real-world applications. As a result, the need for responsible machine learning has emerged, focusing on aligning ML models to ethical and social values, while enhancing their reliability and trustworthiness. Responsible ML involves many issues. This survey addresses four main issues: interpretability, fairness, adversarial robustness, and domain generalization. 
   % Traditional feature selection, a crucial step in ML, plays a significant role in enhancing model performance and interpretability. 
    Feature selection plays a pivotal role in the responsible ML tasks. However, building upon statistical correlations between variables can lead to spurious patterns with biases and compromised performance. This survey focuses on the current study of causal feature selection: what it is and how it can reinforce the four aspects of responsible ML. By identifying features with causal impacts on outcomes and distinguishing causality from correlation, causal feature selection is posited as a unique approach to ensuring ML models to be ethically and socially responsible in high-stakes applications.% Furthermore, this survey includes in-depth insights into how different methods leverage causal feature selection across various real-world datasets.
\end{abstract}

\section{Introduction}
% what is responsible machine learning discuss the tasks: interpretability, fairness, domain generalization and adversarial robustness) you can use material from "causal feature selection in responsible machine learning section
% how does feature selection play a part in responsible machine learning
% why do we need "causal" feature selection
% the contribution of the survey : this is the first effort that categorizes and summarizes efferts on this topic
Machine learning (ML) has been deeply and quickly incorporated into many facets of contemporary life. These days, ML models are used for vital tasks like disaster relief~\cite{berariu2015understanding} and disinformation campaigns~\cite{islam2020deep}, in addition to improving routine tasks like multimedia content recommendations~\cite{sheth2023causal}. But the rapid use of these technologies has also brought to light serious issues with their reliability, transparency, and ethical application~\cite{xu2021machine,rengasamy2021towards}. The discipline of responsible machine learning was founded in response to these worries, with the goal of coordinating ML developments with social, legal, and ethical norms.

Responsible Machine Learning can be defined as an approach that focuses on creating models which not only achieve their intended task but do so in a way that is understandable and transparent, treats all individuals and groups equitably, remains robust against manipulative and deceptive inputs, and performs reliably across a variety of different environments and contexts. This approach ensures that the technology not only serves its immediate purpose but does so in a manner that aligns with broader ethical principles and adapts effectively to diverse and changing conditions~\cite{arrieta2020explainable,hall2020responsible}. For instance, when ML systems are used for hiring processes they might have significant impacts in ensuring responsible decisions. Responsibility of decisions can be quantified across various aspects, including model outputs and model behavior in different situations. When considering model outputs, the responsible ML model should generate (1) interpretable and (2) fair decisions. 
\textit{Interpretability} involves the ML system's ability to provide clear, understandable explanations for its decisions~\cite{carvalho2019machine}. For instance, if a candidate is not selected, the system should be able to articulate why, perhaps due to lacking specific skills relevant to the job. \textit{Fairness} requires that ML models assess candidates equitably, without biases based on gender, ethnicity, or other unrelated characteristics, ensuring all applicants are evaluated solely on their qualifications~\cite{mehrabi2021survey}.

Responsible ML models should also be robust to manipulations and generalizable in unseen scenarios.
\textit{Adversarial Robustness} is about the system’s resilience against manipulative inputs, e.g., when candidates overstate their qualifications~\cite{silva2020opportunities}. This ensures the integrity and reliability of the hiring process. Each of these aspects is crucial in making the ML system effective and ethically sound and trustworthy in its operations. Thus, the responsibility of ML decisions is built upon four key pillars: interpretability, fairness, domain generalization, and adversarial robustness.
\textit{Domain Generalization} refers to the ML's capability to perform accurately across different job roles and contexts, beyond the scenarios it was specifically trained on~\cite{sheth2022domain}. Hiring trends and job market dynamics can change over time. Domain generalization makes sure that ML models do not cling to outdated criteria that could unfairly disadvantage certain groups of candidates.

Feature selection is a fundamental step in the development of ML models, where relevant features are chosen for model training. This process aids in enhancing model performance and computational efficiency. In the context of responsible machine learning, traditional feature selection contributes significantly~\cite{joo2023towards,rastegarpanah2021tools}. By selecting relevant features, models become more interpretable and easier to audit for fairness. It also aids in domain generalization by identifying features that are universally relevant, and in developing adversarial robustness by reducing the model's exposure to misleading inputs. 

Although feature selection has played a significant role in improving the performance of AI models, it mostly concentrates on statistical correlations and may miss the complex dynamics of causality. This becomes especially important in situations where ML choices have significant practical and ethical ramifications~\cite{demirciouglu2021measuring,winkler2019association}. When it comes to recruiting, for example, depending only on correlated features could cause ML systems to base choices on patterns that are only historically common in the data rather than ones that are causally relevant. This could unintentionally reinforce biases or omit important details that are essential to a transparent and successful recruiting process.

This limitation can be addressed with the aid of causal feature selection~\cite{guyon2007causal}. Unlike traditional methods, causal feature selection aims to identify features that have a direct and meaningful impact on outcomes. It differentiates between correlation and causation, ensuring that the model's decisions are based on factors that genuinely influence the results. This approach is pivotal in enhancing the model’s responsibility w.r.t. the different aspects. For example, traditional feature selection might identify a degree from a prestigious university as a key feature because many successful employees have such degrees. However, causal feature selection might reveal that the critical factor is not the prestige of the university but the specific skills and knowledge that the candidate has acquired. This distinction is crucial. By focusing on causal factors, the hiring ML avoids biases associated with certain universities and instead concentrates on the actual competencies required for the job, leading to a more fair and effective hiring process.

This survey aims to close the gap between theoretical goals and practical applications by emphasizing the significance of causal feature selection in responsible machine learning. It emphasizes the role of causality in creating ML systems that are not just technically adept but also morally and socially responsible through a thorough examination of approaches and research in this area. The purpose of the study is to spur more investigation and application in this important field, especially in high-stakes fields where the effects of ML judgments can be profound and far-reaching.

\section{Causal Feature Selection}

Let $X = \{X_{1}, X_{2}, ... X_{N}\}$ be a set of features and the outcome variable be $Y$, represented with a causal graph $G$ where each node is a feature and an edge indicating the causal relationship between the features. Causal feature selection aims to find the Markov Blanket (MB) of the variable $Y$ which includes its direct parents, children and spouses (parents of children) in the causal graph. Given the Markov Blanket $MB$, the outcome variable $Y$, is independent of the remaining features. This implies that the Markov Blanket alone is sufficient to predict $Y$, therefore, providing the causal features required. For a discussion on causal feature selection methods, we direct the readers to \cite{yu2020cfsmethods}.

%The set of causal features comprises of the minimal set that preserves the predictive power over outcome $Y$, without any information about other features.

\begin{figure}
\centering
    \includegraphics[width=13pc]{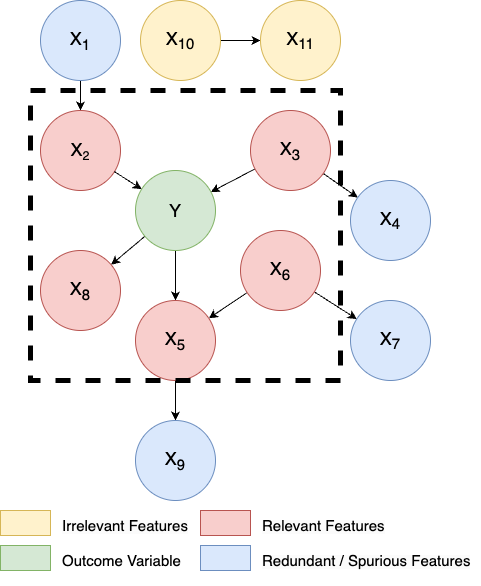}
    \caption{Relevant, irrelevant, redundant/spurious features are annotated in a causal graph.}
    \label{fig:cfs_figure}
\end{figure}

As an illustration, let a causal graph $G$ be defined as shown in Figure \ref{fig:cfs_figure}: we can see that $X_{2}$, $X_{3}$ are direct parents of $Y$ and therefore predictive of the outcome. Similarly $X_{5}$, $X_{8}$ also are predictive of the outcome given that they are direct children of $Y$. We also observe that parents of direct children, like $X_{6}$ is also a causal feature as it gives us information whether it caused the feature $X_{5}$ or if it was purely because of $Y$. In the given causal graph $\{X_{2}, X_{3}, X_{5}, X_{6}, X_{8}\}$ forms the set of relevant causal features to predict outcome $Y$. The set of features $\{X_{1}, X_{4}, X_{7}, X_{9}\}$ is connected to the outcome variable $Y$ though not directly relevant. Hence, they form the set of redundant/spurious features. The set of features $\{X_{10}, X_{11}\}$ is irrelevant given that they are completely disconnected with the outcome variable $Y$.

\section{Causal Feature Selection in Responsible Machine Learning} 

\begin{figure*}
\centering
    \includegraphics[width=\textwidth, height=10pc]{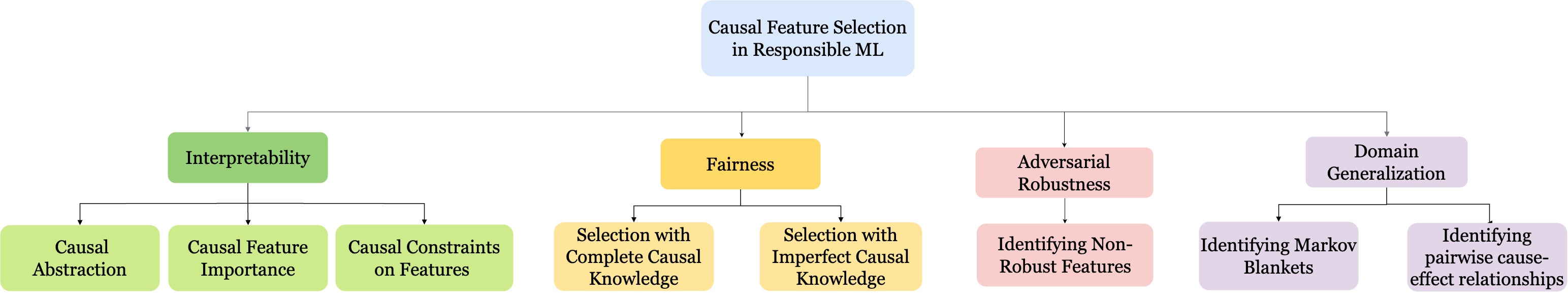}
    \caption{Taxonomy of the Responsible ML tasks in terms of usage of the causal feature selection.}
    \label{fig:tax}
\end{figure*}

In the realm of Responsible Machine Learning, particularly in sensitive applications like hiring, the selection of features plays a pivotal role in ensuring fairness and eliminating bias. This section delves into `Causal Feature Selection', a methodological approach that prioritizes not just the predictive power of features, but also their relevance and appropriateness in causal relationships. By scrutinizing the causal links between features and outcomes, we can develop responsible machine learning models that not only perform effectively but also uphold ethical standards and fairness. This approach is especially crucial in applications such as hiring practices, where reliance on non-causal or spurious correlations could perpetuate biases and lead to unjust outcomes. Through causal feature selection, we aim to foster models that make equitable and justifiable decisions, aligning with the core principles of Responsible Machine Learning. The taxonomy of the Responsible ML tasks in terms of usage of the causal feature selection can be visualized as seen in Figure~\ref{fig:tax}.

\subsection{Interpretability}\label{interpretability}

With the rise of large complex deep learning models, the importance of interpretability to understand the decision making of models has become important. Traditional interpretable methods leverage correlations to identify important features, resulting in obtaining features that merely correlate with the outcome, providing a falsified sense of importance. Causal interpretability avoids this problem by identifying the causal effects on the outputs. Causal feature selection methods can be used in this regard to obtain causal feature importances.

Causal abstraction, as explored in~\cite{beckers2019abstracting} and further discussed by~\cite{geiger2021causal,geiger2023causal}, introduces a method to align explainable, high-level causal models (denoted as $\mathcal{H}$, e.g., tree-based models) with more complex, low-level models (denoted as $\mathcal{L}$, e.g., deep neural networks). The primary goal is to abstract the complex model $\mathcal{L}$ to discover the causal features that underpin it.
To achieve this, the authors map clusters of components from the high-level model $\mathcal{H}$ to each component in the low-level model $\mathcal{L}$. Specifically, let $V_{\mathcal{L}}$ and $V_{\mathcal{H}}$ represent the variables in the low-level and high-level models, respectively. For each variable $X_{\mathcal{H}}$ in the high-level model, they define a partition $\pi_{\mathcal{X}_{H}}$ within $V_{\mathcal{L}}$, excluding certain `empty' variables represented as ${\perp}$, which do not map to any high-level variable. This partition is such that there exists a partial surjective map $\tau_{\mathcal{X}_{H}}: \text{Val}(\pi_{\mathcal{X}_{H}}) \rightarrow \text{Val}(\mathcal{X_{H}})$. This indicates that a group of lower-level variables maps onto a single variable in the higher-level model, with some lower-level variables remaining unmapped.
Furthermore, to find the correct alignment between $\mathcal{L}$ and $\mathcal{H}$, they utilize the concept of `interchange interventions'. This concept ensures that the interventional distributions of both models are equivalent, indicating that changes in the low-level model reflect accurately in the high-level model and vice versa. Through this process, causal abstraction bridges the gap between complex computational models and more interpretable, high-level representations.

The study by~\cite{aditya2019nnattribution} focuses on quantifying the causal effect of an individual input neuron on a specific output neuron within neural networks. This approach is applied to both feed-forward neural networks and recurrent neural networks by transforming them into causal models for analysis. To assess the causal influence, consider an input neuron represented as $x_i$ and an output neuron as $y$. The methodology calculates the average causal effect (ACE) of $x_i$ on $y$. This is achieved by comparing two scenarios: the interventional expectation (where the value of the input neuron $x_i$ is set to a specific value $\alpha$ through an intervention) and a baseline scenario (where there is no intervention on $x_i$). The average causal effect is mathematically expressed as:
\begin{displaymath}
ACE^y_{do(x_i=\alpha)}=\mathbf{E}[y | do(x_i=\alpha) ] - baseline_{x_i}
\end{displaymath}
In this equation, $\mathbf{E}[y | do(x_i=\alpha) ]$ represents the expected value of the output neuron $y$ when the input neuron $x_i$ is intervened upon and set to $\alpha$, and $baseline_{x_i}$ is the expected value of $y$ when there is no intervention on $x_i$. By identifying input features that have a substantial causal effect on the output, the model provides a more nuanced understanding of feature importance, grounded in causal reasoning.

Another approach for identifying causal features in predictive models involves measuring the causal effect of features on model outputs, aiding in feature ranking.~\cite{schwab2019cxplain} applies Granger's concept of causality~\cite{granger69} to train a separate explanation model, aimed at interpreting the importance of features in any prediction model. Let $\hat{f}$ be a predictor, with input features $X$, ground truth $y$, model predictions $\hat{y}$, and a corresponding explanation model $\hat{A}$. The predictive loss is denoted as $\varepsilon_{X}=\mathcal{L}(y,\hat{y}_{X})$.
For a specific feature $X_{i}$, the objective based on Granger's causality is defined as follows:
\begin{displaymath}
\Delta \varepsilon_{X, i} = \varepsilon_{X \setminus {i}} - \varepsilon_{X}
\end{displaymath}
\begin{displaymath}
w_i(X) = \frac{\Delta \varepsilon_{X, i}}{\sum_{j=0}^{p-1} \Delta \varepsilon_{X, j}}
\end{displaymath}
\begin{displaymath}
\mathcal{L}_{granger} = \frac{1}{N} \sum_{l=0}^{l=N-1} KL(\omega_{x_l}, \hat{A}_{x_l}),
\end{displaymath}
where, $\omega$ signifies the distribution of feature importances calculated by excluding each feature from the predictor. In contrast, $\hat{A}$ represents the distribution of feature importances as determined by the explanation model. Minimizing this objective function helps pinpoint the causal effects of various input features on the outputs. Additionally,~\cite{schwab2019granger} presents a framework to quantify the uncertainty in these feature importances. They employ an attentive mixture of experts (AMEs) approach, where each feature is processed by an individual neural network (expert), yielding both a hidden output state and a contribution value that reflects its importance in the AME's prediction. Traditional measures like mutual information gain are insufficient for identifying causal features. The authors of~\cite{panda2021InstancewiseCF} propose the use of a causal metric, Relative Entropy Distance (RED), for instance-wise identification of causal features.

Deep Neural Networks (DNNs) are known for extracting latent features that are often interpretable to humans, as highlighted in~\cite{krizhevsky2012imagenet}. Understanding how these latent features causally affect the predictions of a model can significantly enhance the transparency of the decision-making process. To analyze these causal effects, the Rubin causal model, which hinges on counterfactual reasoning, is commonly used. The essence of this approach is to determine how altering a feature (creating a counterfactual scenario) impacts the model's output.
A key concept in this domain, introduced in~ \cite{goyal2020explaining}, is the Causal Concept Effect (CaCE), which quantifies the causal impact of latent features on a model’s output. For a binary concept $C$, an input $I$, and a classifier $f$, CaCE is mathematically represented as:
\begin{displaymath}
CaCE(C, f) = \mathbf{E}[f(I) | do(C=1)] - \mathbf{E}[f(I) | do(C=0)]
\end{displaymath}
This equation emphasizes the need for counterfactual data points to compute CaCE. However, engineering such interventions in complex datasets with multiple latent features is challenging due to the requirement of a complete causal graph.~\cite{goyal2020explaining} addresses this by employing Variational Auto-Encoders (VAEs) to generate counterfactuals after an intervention.
The authors of~\cite{parafita2019explaining} critiques the limitations of standard counterfactual image generators, which rely on observational data, and suggests a framework for using causal counterfactuals to pinpoint causal features. This involves constructing a causal graph of latent factors and generating counterfactual images from these factors to estimate intervention effects.
In~\cite{dong2022geass} the authors propose a neural-network-based method for extracting causal features, specifically tailored for complex biological data that encompasses large spatio-temporal dimensions. This approach utilizes a modified transfer-entropy loss function to identify causal features effectively.

Counterfactual explanation, as a concept, involves altering a data sample minimally to change the outcome, thereby helping to identify causal feature importances. However, a common issue with these explanations is their lack of consideration for inherent causal relationships. For instance, certain sample features like age, education, or gender cannot be realistically modified. Addressing this~\cite{mahajan2019preserving}, propose a method for generating counterfactual explanations that adhere to both global constraints (which apply universally to all samples) and local constraints (which are specific to individual samples). This is achieved by employing a causal proximity loss to ensure these constraints are respected.
Further expanding on this idea, the authors of~\cite{Kommiya_Mothilal_2021} establish a connection between counterfactual explanations and feature attributions. They utilize methods like~\cite{wachter2017counterfactual} and DiCE~\cite{mothilal2020dice} for generating explanations. They introduce two measures, necessity ($\alpha$) and sufficiency ($\beta$), defined as follows:
\begin{displaymath}
\alpha = Pr(x_{j} \leftarrow a' \Rightarrow y \neq y* | x_{j} = a, x_{-j} = b, y = y*)
\end{displaymath}
\begin{displaymath}
\beta = Pr(y=y* |x_{j} \leftarrow a)
\end{displaymath}
These measures are inspired by the causality framework discussed in~\cite{halpern2016actual} and are used to evaluate feature importances. The authors argue that traditional methods like LIME~\cite{ribeiro2016lime} and SHAP~\cite{lundberg2017shap} may not effectively identify necessary or sufficient features for causal explanations.

In~\cite{karimi2021algorithmic} the authors critically assess the use of counterfactual explanations in decision-making. They highlight a key limitation: these explanations often fail to consider underlying causal relationships and practical feasibility, leading to explanations that are not only sub-optimal but sometimes entirely infeasible. To address this, the authors propose an alternative approach termed minimal intervention'. This method prioritizes actions that are both feasible and plausible within a causal framework, aiming to minimize the cost of implementing changes.
The core idea is to select the optimal set of actions, denoted as $\mathbf{A}^*$, from a set of feasible actions $\mathcal{F}$ and plausible actions $\mathcal{P}$. These actions are chosen to minimize the cost of recourse while ensuring a change in the outcome of the model $f$. The process is mathematically formulated as follows:
\begin{displaymath}
\begin{aligned}
\mathbf{A}^* \in \underset{\mathbf{A}}{\operatorname{argmin}} & \operatorname{cost}\left(\mathbf{A} ; \boldsymbol{x}^{\mathrm{F}}\right) \\
\text { s.t. } & f\left(x^{\mathrm{SCF}}\right) \neq f\left(x^{\mathrm{F}}\right) \\
& x^{\mathrm{SCF}}=\mathbb{F}_{\mathbf{A}}\left(\mathbb{F}^{-1}\left(x^{\mathrm{F}}\right)\right) \\
& x^{\mathrm{SCF}} \in \mathcal{P}, \quad \mathbf{A} \in \mathcal{F}
\end{aligned}
\end{displaymath}

The authors of~\cite{paul2017feature} address the use of causal features for transparent decision-making in machine learning tasks, such as document classification. The proposed method focuses on learning causal relationships between word features and document classes, thereby avoiding the pitfalls of spurious correlations often encountered in traditional approaches. This is achieved by comparing two sets of documents: a treatment group containing a specific word and a control group without that word. The likelihood of word occurrence is quantified using propensity scores, denoted as $P(w|\mathrm{d_{i}}-{w})$. This comparison helps in identifying words that have a causal relationship with the document classes.
To validate the significance of these word-class associations, the method utilizes McNemar's test~\cite{mcnemar1947note}, which is expressed mathematically as:
\begin{equation*}
\chi^2 = \dfrac{(TN-CP)^2}{(TN+CP)},
\end{equation*}
where $TN$ is the count of negative class labels in the treatment group, and $CP$ the count of positive class labels in the control group, to identify significant word-class associations.

Healthcare applications necessitate identifying causal features as spurious features may lead to detrimental results. The paper~\cite{chen2022interpretable} presents an instance-wise causal feature selection framework for developing reliable disease prediction models, particularly important in healthcare to avoid reliance on spurious correlations. This framework utilizes a counterfactual feature selector network alongside factual and counterfactual prediction networks, optimizing prediction errors for both. It also estimates the causal effect of each feature on the output for individual instances.

% One other application is in software debugging. Identifying the root cause of a bug is often very important in software development. In \cite{dubslaff2022causality} causal feature selection is used for root cause analysis of bugs and crashes, and to understand various interactions between features. For a given set of effect properties (like long execution time, crashes) causal features (say signing, encryption) are identified. Notions like blame and feature responsibility are formulated to identify the causal features.  

In software debugging, the authors of~\cite{dubslaff2022causality} introduce causal feature selection for root cause analysis of bugs and crashes. This approach identifies causal features (e.g., signing, encryption) responsible for certain effects (like long execution time, crashes) and formulates concepts of blame and feature responsibility.
%\saketh{=================================}
%\saketh{adding papers from other sections (from Prediction/Classification, misclassified fairness paper) below!}
Most of the feature selection methods identify the same set of features for different labels. In multilabel datasets the interactions are among the labels, features, and between labels and the features. The goal in \cite{wu2020multi} is to identify unique causal features for different labels with a Markov Blanket based multilabel causal feature selection. The algorithm proceeds in three phases. Firstly the local causal structure is determined for each of the labels. Then, the features ignored by strong label relevance are retrieved. Finally, the common features affecting the labels and label-specific features are extracted. \cite{aliferis2010local} propose Generalized Local Learning (GLL), a framework for identifying the causal local structure of the target variable for the classification task. This framework, which is built upon conditional independence tests is shown to be sound under sufficient conditions. %\raha{more causal discovery for feature selection rather than application of causal feature selection}

\subsection{Fairness}

Developing fair models without discrimination against any specific group is crucial to ensure equitable treatment and uphold ethical principles in decision-making for real-world applications. Statistical notions of fairness select features that are implicitly affected by sensitive features such as race and gender, resulting in unfair biased decision~\cite{makhlouf2020survey}.
In contrast, causal notions of fairness utilize causal feature selection to assess and remove the causal impact of sensitive features on the models' outcome. 
% Based on our knowledge of the causal graph we can categorize the approaches into two categories: fairness with complete causal graph, and fairness with imperfect causal graphs.

% To understand the fairness of the outcomes given by the machine learning models, it is important to understand the process behind the decision-making. Using purely statistical frameworks, we cannot account for the bias of the sensitive features on the outcomes of a machine learning model. Causal features can contribute in this regard making the process by separating this bias and also making the process more transparent.

% \subsubsection{With Complete Causal Graph}

Numerous causal fairness methods have been developed with the assumption of full knowledge of the causal graph of demonstrating the causal relations between the variables in the system. \cite{kusner2017counterfactual} introduced counterfactual fairness to select causal features for building fair predictors. Let $A$, $X$, $Y$ be the set of admissible attributes, normal attributes (attributes other than admissible) and the outcome, counterfactual fairness is given as follows: 
\begin{displaymath}
    P\left(\hat{Y}_{A \leftarrow a}(U)=y \mid X=x, A=a\right) = 
\end{displaymath}
\begin{displaymath}    
    P\left(\hat{Y}_{A \leftarrow a^{\prime}}(U)=y \mid X=x, A=a\right)    
\end{displaymath}

Intuitively, if the outcome probability of a model for a given individual instance, is same in the observational scenario, and in the counterfactual scenario where it takes different attribute value for the sensitive variable, we call the model counterfactually fair. The paper outlines three levels of assumptions that ensure that the selected features result in a counterfactually fair predictor. Firstly, selecting only non-descendants of sensitive features $A$ ensures counterfactual fairness, though most of the features often descend these sensitive features in the causal graph. Secondly, the information about $X$ is passed to the model from the posterior distribution $P(U | X, A)$ where $U$ represents latent unobserved variables that are learnt. Thirdly, a fully deterministic causal model can be learnt to build a counterfactually fair predictor.

%formula
% In \cite{kusner2017counterfactual,loftus2018causal} the authors discuss how various notions of fairness fail to address discrimination and propose counterfactual fairness. The idea is that if the predictor does not flip its decision on changing the protected attributes, keeping the remaining attributes constant, it is considered a fair predictor. Causal features that are non-descendants of admissible attributes in the causal graph are utilized to achieve counterfactual fairness. Let A, X, Y be the admissible, normal (attributes other than admissible) and outcomes, the definition of counterfactual fairness is as follows:

% \begin{displaymath}
%     P\left(\hat{Y}_{A \leftarrow a}(U)=y \mid X=x, A=a\right) = 
% \end{displaymath}
% \begin{displaymath}    
%     P\left(\hat{Y}_{A \leftarrow a^{\prime}}(U)=y \mid X=x, A=a\right)    
% \end{displaymath}
    
% The authors propose a fair-learning approach, in which any model is trained with the causal non-descendant features of protected attributes. Each data point in the dataset is further augmented by adding unobserved attributes sampled using the Markov chain Monte Carlo method. 

Sometimes selecting causal features which are discriminatory (e.g., race, gender) can perpetuate biases through other proxy variables (e.g., their hobbies) resulting in proxy discrimination. In \cite{kilbertus2017avoiding} a detailed framework for avoiding proxy discrimination is provided with the assumption of having the complete causal graph. The idea of resolving variables is also provided, through which sensitive features are allowed to affect the final prediction. A framework to eliminate unresolved and proxy discrimination is given along with the assumptions made on the causal model for the formulation to work.

% \saketh{exactly same papers, remove one of the two below!}
% The goal of \cite{singh2019fair} is to maintain fairness constraints under different distributions encountered during test/deployment settings. Specifically the task is to find predictive features in the target domain, by utilizing labelled data from the source domain without deprecating the fairness. Most of the previous works concern maintaining predictive accuracy under distributional shifts. Though none of them focus on maintaining fairness under the shifts. The paper makes two assumptions i.e. 1) given the feature subset used for the fair predictor, domain is independent of the outcome and 2) feature subset is independent of the domain given the outcome and domain. Then they formulate an approach that generalizes to distribution shifts under these assumptions. The approach tries to find features sets in increasing order of source domain empirical risk \saketh{define this??}  and then tries to select features satisfying the required assumptions by using notions of d-separation in the causal graph. 

Using only causal features for domain generalization with access to some unlabeled samples from target domain(s) may be detrimental to the fairness under distribution shifts. The problem of maintaining fairness under distribution shifts is discussed in~\cite{singh2021fairness}. Any model seeking to reduce the effect of distribution shifts without considering the uneven distribution of the training data will likely fail to mitigate the effects of distribution shifts on fairness. The paper proposes choosing a set of features given which the domain and outcome are independent. These sets of features form the separating set as they d-separate domain and outcome in the causal graph when conditioned upon. The paper also postulates that unless a few assumptions on the invariances of classification error and fairness constraints hold, this problem of fair generalization cannot be solved. The fair-learning approach suggested first ranks all the feature subsets based on their errors in source domains. It then selects a subset from these features moving in the sorted order, to find a subset that satisfies the invariance assumptions. This is shown to perform better than using only the causal features for the relaxed case of domain generalization based on both accuracy and fairness metrics.

\cite{ling2023fair} propose a $K$-Fairness algorithm for context $K$, a set of sensitive features $S$, outcome variable $Y$  as:

\begin{displaymath}
    P(Y=y | do(S=0, K=k)) = P(Y=y, do(S=1, K=k))
\end{displaymath}
Based on this notion of fairness \emph{FairCFS} algorithm proceeds by firstly identifying the Markov Blankets of $Y$, $S$ given by $MB_{Y}$ and $MB_{S}$. Then, all the features independent of S from the set of features $MB_{Y} \setminus MB_{S}$ are identified comprising set of features $M_{1}$. Finally,  features independent of $S$ from $MB_{Y} \cap MB_{S}$ indicated by feature set $M_{2}$ is found. $M_{1} \cup M_{2}$ provide causally fair features for the outcome $Y$.

Relaxing the assumption of full  causal graph knowledge, recently a series of methods with requirement of partial knowledge of the causal graph have been developed. Conventional feature selection algorithms exploit correlations and fail to protect the bias of sensitive features on the outcomes. In \cite{galhotra2022causal}, causal feature selection is used for mitigating bias in data integration pipelines, which mostly refers to adding new features to a dataset. The paper discusses the notion of admissible attributes, a set of features through which it is permissible for sensitive features to affect the target outcome. Let $A$, $S$, and $Y$ be admissible, sensitive features and the outcome variable. A group testing algorithm is proposed to identify the candidate features to be added. In the first phase, a set of variables $C_{1}$ independent of sensitive features in the presence of admissible variables i.e. $X \perp S | A$ are chosen. In the second phase, the subset of variables $C_{2}$ that do not affect target variables are found i.e. $X \perp Y | A$, forming $C_{2}$. $C_{1} \cup C_{2}$ form the final causal features selected. This work does not demand a complete causal graph apriori unlike other works.
%While most of the causal approaches demand a complete causal graph apriori, there is no such requirement in this work.
Usually identifying the true causal graph uniquely from the observational data is not possible. \cite{zuo2022counterfactual} guides in selecting the causal features to achieve counterfactual fairness in settings where only the partial causal graph is known. MPDAGs (Maximal Partially Directed Graphs) can be learned, which essentially includes information about all the sets of Markov equivalent PDAGs. A given predictor is fair, if descendants of sensitive features from the MPDAG are not used as features. The paper introduces a few lemmas to identify ancestral relationships between pairs of variables from the MPDAG, identifying whether one variable is a definite descendent of the other. On many benchmarks, \textit{FairRelax}, where all definite non-descendants of sensitive features and some possible descendants of sensitive features are used, outperforms others when evaluated on counterfactual fairness metric.

\subsection{Adversarial Robustness}
%\saketh{Correct all the references. Also how is taxonomy application 1 different from taxonomy application 2 (ask Raha)?}

Despite the tremendous success of Deep Neural Networks (DNNs), studies have exposed their vulnerability to adversarial attacks, which craft adversarial examples with human-imperceptible perturbations that lead to erroneous predictions by these models~\cite{szegedy2013intriguing,goodfellow2014explaining}. 
%%%%%%% Use features selection to explain why adversarial examples exist
The primary application of causal feature selection is to elucidate the reasons for the presence of adversarial examples by pinpointing the features responsible for the model's susceptibility to adversarial perturbations. \cite{ilyas2019adversarial} demonstrate that the vulnerability of DNNs to adversarial examples is attributed to the existence of non-robust features, a.k.a. spurious features, that are highly predictive but are incomprehensible to humans. A series of methods are then developed to identify and select such feature.
\cite{ilyas2019adversarial} propose a theoretical framework whcih categorizes the features into three main types: useful features, robustly useful features, and useful, non-robust features. The usefulness of features is measured based on their correlation with the label. A set of features are called robust if after being perturbed they are still correlated with the label. They then propose an approach to disentangle robust features from non-robust ones and construct a robust dataset that consists of only robust features and a non-robust dataset with only non-robust features. Specifically, to capture only the robust features in input $x_r$, its features are forced to be similar to the ones extracted by a robust classifier:

\begin{equation*}
%\begin{aligned}
\label{eq:robust}
\min_{x_r}||g(x_r)-g(x)||_2,
\end{equation*}
where $g(.)$ is the output of the penultimate layer of a robust (i.e., adversarially trained) neural network.

To extract non-robust features of an input $x_r$, a target class $t$ is selected randomly or deterministically according to the source class. An adversarial example of $x_r$ is then generated. The non-robust features of this adversarial example are non-robust whereas the robust ones are the robust features of the true groundtruth label.
 \cite{singla2022salient} propose a general framework to discover a subset of spurious and core visual features used by deep models. to identify spurious or core visual attributes used for predicting the class $i$, they define core visual features and spurious features as follows: A core visual feature is an attribute for class $i$ that is always a part of the object defined by the class. A feature is spurious if not it is not core. To annotate a neural feature as core or spurious, they adopt only the top-5 images (with a predicted label of $i$) that maximally activate that neural feature and use Mechanical Turk workers to annotate. They then show that these neural feature annotations generalize extremely well to top-k images with label $i$ that maximally activate that neural feature. 
They then utilize their method to generate the Salient ImageNet dataset that contains core and spurious masks for a huge fraction of samples from the  ImageNet.
\cite{jha2019attribution} utilize ideas from neural network attribution (explained in Sec.~\ref{interpretability}) to identify such features. The main idea of this work is that adversarial examples exist because of a relatively small
number of features with high attribution in the machine learning model. Specifically, the paper shows that it is possible to identify an input is non-robust and has an adversarial example by examining inputs in its
causal neighborhood obtained by incrementally masking the features which have high magnitude attributions. 

\subsection{Domain Generalization}

In machine learning, domain generalization (DG) is a pivotal concept for enhancing model performance on unseen data. 
% DA differs from DG in that it accesses some unlabeled samples of the test data, offering a glimpse into the new domain. 
An effective strategy for achieving domain generalization is through causal techniques that focus on learning invariant features~\cite{sheth2022domain}. These techniques, particularly causal feature selection methods, prioritize features causally linked to the target variable over mere correlations. This approach enhances model's functionality against data distribution changes, as causal features tend to remain stable across varied domains.

% In machine learning, domain adaptation or domain generalization refers to the ability of a model to perform well on previously unseen data without being trained on each new domain. The key difference between domain adaptation (DA) and domain generalization (DG) is the fact that, in DA the model has access to some of the unlabeled samples of the unseen test data. One way to achieve domain generalization is to leverage different causal techniques to learn the invariant features ~\cite{sheth2022domain}. One such group of causal methods is causal feature selection methods. These methods aid in selecting features that are causally related to the target variable rather than simply correlating with it. This ensures that the model is robust to changes in the data distribution, as the causal features are more likely to be stable across different domains. In this section, we discuss works that leverage causal feature selection for domain generalization.

% Domain adaptation and generalization methods aim to improve deep learning models' robustness and generalizability. A series of works showed that relying on causal features aid in capturing the invariance in the data across various environments~\cite{sheth2022domain}. Thus, leading to better generalization. As a result, recent works focus on discovering causal relations for better generalization.

% \raha{be consistent Markov Blanket or Markov Blanket}
% \raha{Markov Blanket is a reoccurring term so it's better to define it once int he preliminary and refer to it multiple times}

Our survey categorizes the majority of causal feature selection methods for DG. The first category includes methods utilizing Markov Blankets for identifying invariant causal features. Markov Blankets form a subgraph around a target node containing all necessary information for predicting the target's behavior. Another category encompasses methods that discover pairwise cause-effect relationships, employing techniques like meta-learning, confounder balancing, and assessing the invariance of causal structures.

%One challenge in this field is obtaining confidence intervals for causal discovery (CD) methods, which require understanding the distribution of causal effect estimators post graphical structure estimation. Addressing this
~

To identify the Markov Blanket, \cite{peters2015causal} propose a method that exploits the conditional distribution of a target given its direct causes, which remains unaffected by interventions on other variables. This method first identifies statistically invariant causal predictors across environments. Among these, the causal submodel—directly influencing the target variable—maintains invariance with a controlled high probability, minimizing false causal discoveries and providing confidence intervals for causal structure.

\cite{yu2019multi} highlight a limitation in current causal feature selection: many methods are not adaptable to multiple datasets addressing the same problem. They introduce a multi-source causal feature selection approach based on ``causal invariance'' The goal is to find a subset of features $S$ where the conditional distribution of target $T$, given $S$, remains stable across different domains $D=\{D_1, D_2, \ldots, D_K\}$. This stability is formalized as:

\begin{equation}
\begin{aligned}
S^*= & \arg\max _{S \subseteq MB(T)} P^i(T \mid S) \\
& \text { s.t. } P^i(T \mid S)=P^j(T \mid S)(\forall j, j \neq i) .
\end{aligned}
\end{equation}

\noindent where $MB(T)$ represents the Markov Blanket w.r.t. the target variable $T$, $i$ and $j$ represent different domains. By selecting $S$ from $MB(T)$, this method identifies invariant causal parents of 
$T$, enhancing generalization in the presence of distribution shifts. In a similar vein, \cite{javidian2021scalable} propose a Scalable Causal Transfer Learning (SCTL) approach for a more relaxed variant of generalization, where the model has access to unlabeled samples from the target domains. SCTL searches the Markov Blanket of target variable $T$ for causally invariant features, addressing scalability issues and aiming to find a separating set $S \subset V$  where $V$ is the set of context variables $C_{i \in I}$ and system variables $X_{j \in J}$ such that for each target variable $T$ the condition $C_i \perp T \mid S$ is satisfied, for every $i \in I$ in the source domain. 

To identify the causal features through pairwise cause-effect relationships, \cite{shen2018causally} integrates confounder balancing with weighted logistic regression, enhancing the model's generalization capabilities. Their method employs a causal regularizer, treating each feature as a treatment variable and optimizing sample weights for balanced distribution in treated and control groups. The objective function for their causally regularized logistic regression model is:

\begin{equation}
\begin{array}{ll}
\min & \sum_{i=1}^n W_i \cdot \log \left(1+\exp \left(\left(1-2 Y_i\right) \cdot\left(x_i \beta\right)\right)\right), \\
\text { s.t. } & \sum_{j=1}^p\left\|\frac{X_{-j}^T \cdot\left(W \odot I_j\right)}{W^T \cdot I_j}-\frac{X_{-j}^T \cdot\left(W \odot\left(1-I_j\right)\right)}{W^T \cdot\left(1-I_j\right)}\right\|_2^2 \leq \gamma_1, \\
& W \geq 0,\|W\|_2^2 \leq \gamma_2,\|\beta\|_2^2 \leq \gamma_3, \quad\|\beta\|_1 \leq \gamma_4, \\
& \left(\sum_{k=1}^n W_k-1\right)^2 \leq \gamma_5,
\end{array}
\end{equation}

\noindent where $\sum_{i=1}^n W_i \cdot \log \left(1+\exp \left(\left(1-2 Y_i\right) \cdot\left(x_i \beta\right)\right)\right)$ represents the weighted logistic loss and the constraints $\|\beta\|_2^2 \leq \gamma_3$ and $\|\beta\|_1 \leq \gamma_4$ help avoid overfitting. The term $W \geq 0$ constrains each of sample weights to be non-negative. The formula $\left(\sum_{k=1}^n W_k-1\right)^2 \leq \gamma_5$ avoids all the sample weights to be 0. $W$ is the sample weights. $\left\|\frac{X_{-j}^T \cdot\left(W \odot I_j\right)}{W^T \cdot I_j}-\frac{X_{-j}^T \cdot\left(W \odot\left(1-I_j\right)\right)}{W^T \cdot\left(1-I_j\right)}\right\|_2^2$ represents the loss of confounder balancing when setting feature $j$ as a treatment variable, and $X_{-j}$ is all the remaining features (i.e. confounders). $I_j$ is the $j^{th}$ column of $I$, and $I_{ij}$ refers to the treatment status of unit $i$ when feature $j$ is treatment variable. 

Similarly, the authors of~\cite{kyono2021selecting} argue that traditional models are not applicable for cross-domain treatment effect estimation as they do not account for missing counterfactuals and fail to factor in the model's predictions in the target domain. To address this problem, the authors propose to leverage the invariance of causal structures across domains and introduce a novel metric for model selection designed for treatment effect estimation models. The proposed model leverages the estimated outcomes under different treatment settings on the target domain by incorporating a measurement of how well the outcomes satisfy the causal relationships in the interventional causal graph. The authors coin this measure as causal risk and utilize a log-likelihood function for quantifying the model’s fitness to the underlying causal graph, while providing theoretical justifications for using the causal risk for model selection.

The authors of~\cite{yuan2021meta} propose a meta-learning causal feature selection by addressing the general Non-I.I.D. image classification problem to obtain stable causal features in changing environments. They select causal features and remove spurious features to obtain stable representations for different distributional environments. Also, they optimize causal and network parameters jointly. Similarly, the authors of~\cite{gerhardus_2023} aim to learn the cause-effect relationships between different variables within a specified time range. They further utilize the causal relationships to enhance generalizable prediction of tropical cyclones.

% This overview reveals the diverse and evolving nature of causal feature selection in domain generalization. These approaches, grounded in causal inference, provide a robust framework for enhancing model performance across different domains and datasets, a critical aspect in the ever-changing landscape of machine learning.

\section{Conclusion}
In this survey, we discuss the notion of causal feature selection and its role in responsible ML, and shed light on the application of causal feature selection in responsible machine learning — an essential yet emerging topic. We introduce a new taxonomy focused on causal feature selection for the four key responsible machine learning tasks: interpretability, fairness, adversarial robustness, and domain generalization. Representative algorithms are summarized for each task. 
With the help of the taxonomy, we pinpoint the future research directions for each task, for instance, for the \textit{interpretability}, scalable causal feature selection that are compatible with high-dimensional machine learning models methods are required to be developed. In the context of \textit{fairness}, all current methods presume the possession of causal graph knowledge, which proves to be restrictive in real-world applications. This prompts the need for developing methods without prior causal knowledge. 

Causal feature selection is currently served as a method to explain the existence of \textit{adversarial attacks}. For a future direction, causal feature selection methods and their role in developing robust models against such attacks need to be explored. For
\textit{domain generalization}, while current methods focus on identifying pairwise cause-effect relationships or the Markov Blanket w.r.t. a target variable, future research could aim to enhance these methods by incorporating multimodal data sources and temporal dynamics. This could involve developing algorithms capable of handling complex interactions and feedback loops within the data, providing a more nuanced understanding of the causal structure. By delving into the potential of this field, we open a new avenue for the development and utilization of causal feature selection techniques that are suitable for responsible ML.
This catalyzes a shift towards more ethical, transparent, and robust AI, facilitating principled, and groundbreaking advancements in ML.

\newpage
\bibliographystyle{named}
\bibliography{ijcai24}

\end{document}